# A Study on the Matching Rate of Dance Movements Using 2D Skeleton Detection and 3D Pose Estimation: Why Is SEVENTEEN's Performance So Bita-Zoroi (Perfectly Synchronized)?

Atsushi "Woozi pen" Shimojo[1] and Harumi Haraguchi[2*]

[1]Department of Information Engineering, Faculty of Engineering, Ibaraki University, Hitachi, JAPAN
[2]Graduation School of Science and Engineering, Ibaraki University, Hitachi, JAPAN
[*]corresponding harumi.haraguchi.ie@vc.ibaraki.ac.jp(1st author's supervisor)

**ABSTRACT**

SEVENTEEN is a K-pop group with a large number of members 13 in total and the significant physical disparity between the tallest and shortest members among K-pop groups. However, despite their large numbers and physical differences, their dance performances exhibit unparalleled unity in the K-pop industry. According to one theory, their dance synchronization rate is said to be 90% or even 97%. However, there is little concrete data to substantiate this synchronization rate. In this study, we analyzed SEVENTEEN's dance performances using videos available on YouTube. We applied 2D skeleton detection and 3D pose estimation to evaluate joint angles, body part movements, and jumping and crouching motions to investigate the factors contributing to their performance unity. The analysis revealed exceptionally high consistency in the movement direction of body parts, as well as in the ankle and head positions during jumping movements and the head position during crouching movements. These findings suggested that SEVENTEEN's high synchronization rate can be attributed to the consistency of movement direction and the synchronization of ankle and head heights during jumping and crouching movements.

## 1 Introduction

SEVENTEEN(PLEDIS Entertainment) is a male K-pop group that debuted in 2015[1]. Even before their debut, they were known for creating all of their own songs while also delivering outstanding dance performances. With 13 members, they are among the largest K-pop groups, and their physical differences are notable—there is a height gap of over 20 cm between the tallest and shortest members. Additionally, some members have muscular, well-built physiques, while others are so slender that their muscles are barely visible. Despite these significant physical differences, many CARATs (SEVENTEEN's fandom name) actively praise their dance performances-especially the perfection of their Kal-gunmu (perfectly synchronized group choreography)—on social media and video platforms. Even professional dancers in Japan have analyzed their performances on YouTube, expressing astonishment at their high skill level. They have also described SEVENTEEN's exceptional synchronization using the phrase "Bita-zoroi" (Perfectly Synchronized) [2]. Some claim that their synchronization rate reaches 90% or even 97%. However, there are few studies that clearly define the methods used to analyze their dance synchronization. Research on this topic may exist in Korean-language papers, but unfortunately, the author is unable to read Korean and access such insights.

On the other hand, there are several previous studies on dance movements. Qiu, Z et al. (2023) proposed a deep learning-based framework that transforms amateur dance into professional dance to improve dance technique [3]. They utilized three key features of dance movements: key poses such as stops and rotations, spatial variations such as arm swings and jump heights, and temporal variations in music and timing. By leveraging these features, their framework enables the reproduction of visually appealing professional-like dance movements while maintaining the natural flow of motion. Furuichi et al. (2021) generated movement features based on angles and displacement from skeletal data obtained using Kinect to analyze hip-hop dance movements. They demonstrated that individual dances could be accurately identified by focusing mainly on angle-based features through machine learning [4]. They highlighted dance-specific movements such as lifting the legs, jumping, crouching, and chest pops, which involve large whole-body motions and fine stepwork. Their study revealed that factors such as the bending of arms and knees and the degree of displacement contribute to individual identification.

About research on motion similarity, Wenbin X. et al., (2022) studied the control of humanoid robots that accurately mimic human movements in real-time and evaluated the similarity of these movements [5]. Using motion capture, they assessed the similarity between human joint motion trajectories and the corresponding robot trajectories using Dynamic Time Warping (DTW). They demonstrated that DTW provides a more precise method for quantifying trajectory similarity compared to Euclidean distance. However, they identified a limitation: DTW similarity scores only allow for relative comparisons and lack an absolute numerical standard. For example, while it is possible to evaluate how much one trajectory resembles another, interpreting the numerical meaning of the score remains difficult. Zielinska, T. et al. (2023) evaluated the similarity of body part coordination based on human joint trajectories captured using motion capture [6]. They analyzed body part coordination during forward and backward stepping from an upright posture, as well as balance adjustments when pushed from the side. Instead of simply comparing trajectory distances, they used Pearson correlation with the center of mass as a reference, demonstrating that this approach more accurately reflects the characteristics of movement.

There has been research into motion estimation using skeletal detection for some time. The relevant research for this study is Cao et al. (2017) proposed Open Pose, a skeleton detection method using Part Affinity Fields (PAFs), enabling real-time motion analysis for multiple individuals [7]. PAFs learn which joints belong to which individual. Unlike traditional methods that connect detected body parts to joints, this approach learns the relationships between joints. Additionally, Open Pose adopts a bottom-up approach, detecting all joints from the entire image and grouping them by the individual, demonstrating its capability to accurately estimate 2D poses in real-time for multiple people within an image. Do-Hyun K et al., (2022) proposed a method for analyzing worker postures in real-time using 3D pose estimation, automatically detecting the risk of musculoskeletal disorders [8]. Using multiple cameras to detect joint positions, they extracted features such as joint angles, motion trajectories, velocity, and acceleration. This approach enabled the evaluation of static postures and dynamic work-related strain. For worker localization, YOLOv3 was utilized, while 3D pose estimation from 2D skeletons was performed using 3DMPPE.

## 2 Related Methods
### 2.1 YOLO v8

One method for detecting 2D skeletal data of individuals in videos is the YOLO series, an object detection model developed by Ultralytics [9][10][11]. This method can obtain the x and y coordinates of 17 key points on a person in a video, including both eyes, both ears, the nose, both shoulders, both elbows, both wrists, both hips, both knees, and both ankles.

In addition to object detection, the YOLO series also features object tracking, assigning a unique ID to each object in the video to enable continuous tracking across frames. In this study, we utilize YOLOv8, a model released by Ultralytics in January 2023. By combining object detection and object tracking, we can acquire continuous data linked to a specific individual.

### 2.2 Media Pipe

Google's MediaPipe is a well-known framework for estimating 3D poses from 2D skeletal data [12]. MediaPipe is an open-source framework designed to enable real-time, high-precision motion analysis. It provides functionalities for various tasks, including motion analysis, pose estimation, and facial recognition. Its efficient processing pipeline allows lightweight and real-time processing.

The pose estimation module of Media Pipe (Media Pipe Pose) extracts 2D skeletal data from images and videos and estimates 3D poses based on this information. Similar to YOLOv8, it captures the coordinates of key points such as elbows and knees from video frames and constructs a skeletal model based on these coordinates. Then, by applying camera calibration and depth estimation algorithms to the detected 2D coordinates, it projects them into a 3D space.

Figure 1 and Figure 2 present examples of input video frames and the plotted results in 3D space using MediaPipe's 3D pose estimation [13].

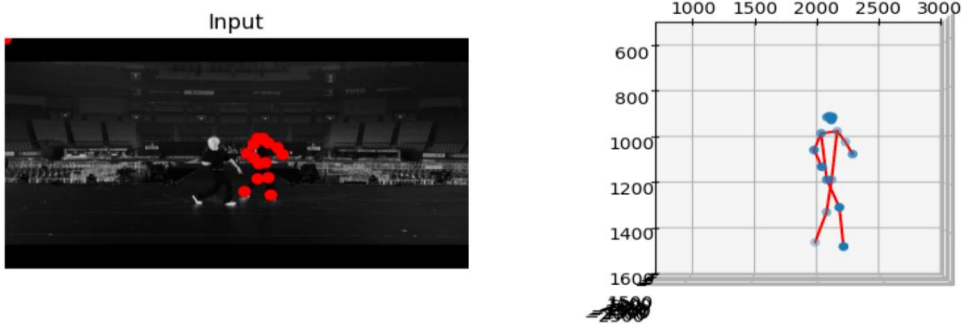

Figure. 2 Example of a 3D posture estimation plot (Front Side)

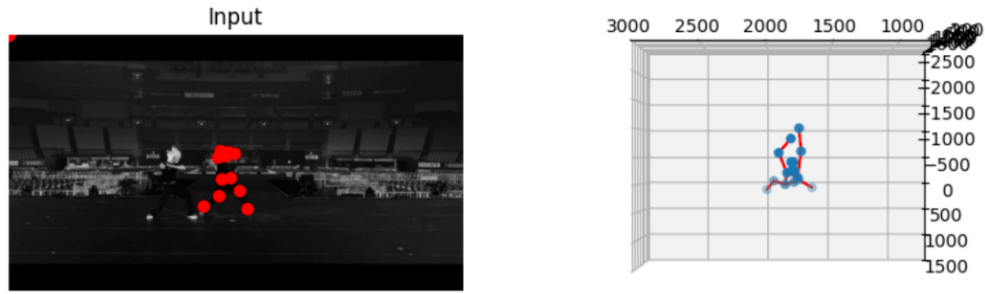

Figure. 1 Example of a 3D posture estimation plot (Above Side)

### 2.3 Dynamic Time Warping(DTW)

Dynamic Time Warping (DTW) is a method for computing the similarity between time series data. DTW computes similarity by constructing a cost matrix and searching for the path with the lowest cumulative cost, dynamically adjusting the optimal mapping on the time axis. The similarity can be calculated even if the length and period of the series are different.

Two time series data $X = (x1, x2, \ldots, xn)$, $Y = (y1, y2, \ldots, ym)$, and $d(xi, yj)$ as distance between each element, the cumulative distance matrix $D(i, j)$ is defined by Equation (1).

$$D(i,j) = D(i,j) = d(xi, yj) + \min\{D(i-1,j), D(i, j-1), D(i-1, j-1)\} \quad (1)$$

However, the boundary condition as follows:

$D(1, 1) = d(x1, y1)$,

$D(i, 1) = d(xi, y1) + D(i-1, 1)$, for $i = 2, 3, \ldots, n$

$D(1, j) = d(x1, yj) + D(1, j-1)$, for $j = 2, 3, \ldots, m$

Finally, the DTW distance between time series data $X$ and $Y$ is given by Equation (2).

$$\text{DTW}(X, Y) = D(N, M) \quad (2)$$

### 2.4 DTW Barycenter Averaging(DBA)

A method for computing an average time series (representative time series) from multiple time series data is DTW Barycenter Averaging (DBA). First, one time series C is selected as the initial value from the given time series. Next, for each time series, the optimal alignment with C is computed using Dynamic Time Warping (DTW). Then, the average values of the corresponding points are updated to derive a new C. This process is repeated until convergence. Given a time series $X_i$, the representative time series C is expressed as Equation (3).

$$C = \arg \min_{C} \sum_{i=1}^{N} \text{DTW}(B, X_i) \quad (3)$$

## 2.5 Cosine similarity

One method for calculating the similarity between two vectors is cosine similarity. Cosine similarity is expressed as Equation (4), where θ is the angle between two vectors, *a* and *b*. Its value ranges from -1 to 1, with higher similarity resulting in a value closer to 1 and lower similarity leading to a value closer to -1.

$$CosineSimilarity = \cos\theta = \frac{\vec{a}\cdot\vec{b}}{\|\vec{a}\|\|\vec{b}\|} \qquad (4)$$

# 3 Analysis Methods

Chapter 3: Overview of Dance Motion Similarity Analysis and Synchronization Score Calculation.

## 3.1 Overview of the Analysis Method

In this study, we first use YOLOv8 to perform 2D skeleton detection and object tracking on dance videos of SEVENTEEN, obtaining bounding box coordinates for each member. Next, using the obtained 2D skeleton data and bounding box coordinates, we extract individual members and apply Media Pipe for 3D pose estimation. Finally, we analyze the time-series data generated from the 3D pose estimation results to evaluate the degree of motion similarity.

## 3.2 Creation of Skeleton Data
### 3.2.1 Detection of 2D Skeleton Data

YOLOv8 is used for 2D skeleton detection. First, the dance video is input, and the bounding box coordinates of each member are obtained. Next, the bounding box coordinates are used to identify individual members, and the x and y coordinates of 17 key points—both eyes, both ears, nose, both shoulders, both elbows, both wrists, both hips, both knees, and both ankles—are extracted.

By combining YOLOv8's 2D skeleton detection with object tracking, we can detect continuous 2D skeleton data associated with each member and their movement range within the video. As an example, Figure 3 and Figure 4 show the first and final frames of the output video from YOLOv8 [14].

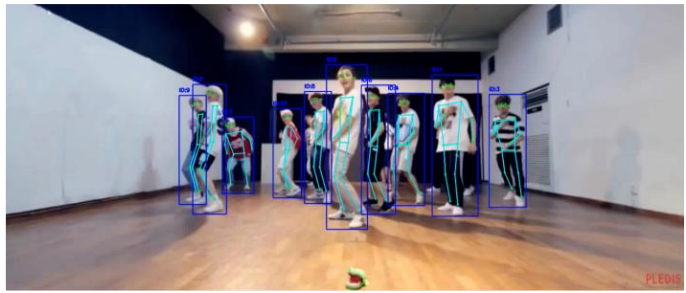

Figure. 3 1st Frame of video

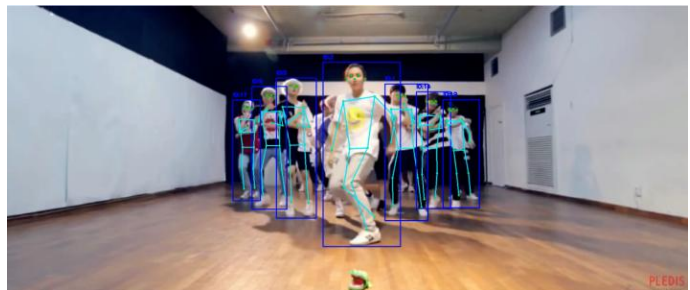

Figure. 4 Last Frame of video

### 3.2.2 3D posture estimation

MediaPipe is used for 3D pose estimation. By inputting the obtained 2D skeleton data and bounding box coordinates into MediaPipe, performing 3D pose estimation for each member became possible. Examples of the plotted results of the 3D pose estimation are shown in Figure 5 and Figure 6.

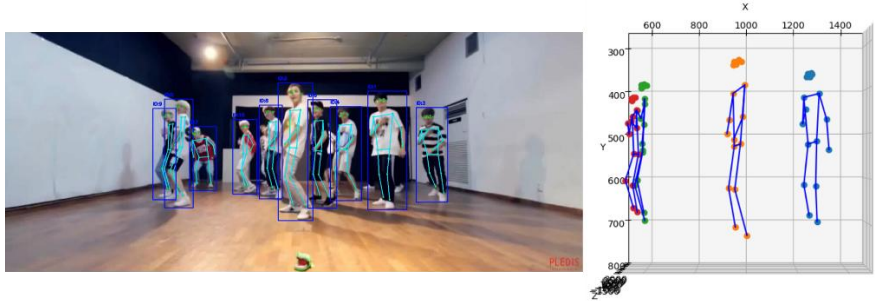

Figure. 5 Plot results for the first frame of video

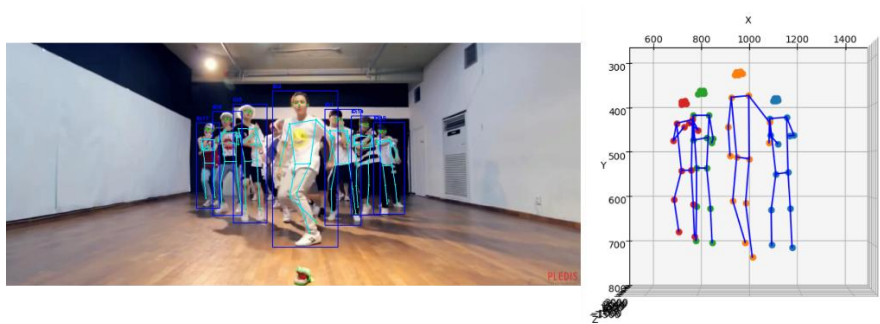

Figure. 6 Plot results for the last frame of video

## 4  Video Analysis

In this chapter, we describe the analysis data, evaluate the similarity of joint angles and motion direction of body parts, and conduct an analysis focused specifically on jumping and crouching movements. The results are then presented.

### 4.1  Overview of the Analysis Method

In this study, we selected scenes from SEVENTEEN's dance videos featuring the "Kal-gunmu" performance. Kal-gunmu is a K-POP dance style characterized by sharp, synchronized movements performed by all members.

From the Kal-gunmu videos, we extracted:

- 20 scenes (dance-scene1 to dance-scene20) featuring 3 to 5 seconds of synchronized choreography.
- 5 scenes (jump-scene1 to jump-scene5) containing jumping movements and their surrounding frames.
- 5 scenes (down-scene1 to down-scene5) featuring crouching movements and their surrounding frames.

For each scene, we analyzed four front-row members as the target subjects. The frame rate was set to 24 fps. The three types of movements mentioned above were analyzed separately.

### 4.2  Evaluation Metrics and Results
#### 4.2.1 Evaluation Based on Joint Angle Similarity

For dance-scene1 to dance-scene20, we assessed the consistency of joint angles throughout the scenes based on 3D pose estimation results. The target joints for evaluation were six joints: both elbows, both knees, and both shoulders (underarms).The changes in these six joint angles were converted into time-series data per frame, and their similarity was evaluated using the DTW-based synchronization score. For dance-scene1 to dance-scene20, we assessed the consistency of joint angles throughout the scenes based on 3D pose estimation results. The target joints for evaluation were six joints: both elbows, both knees, and both shoulders (underarms).

The changes in these six joint angles were converted into time-series data per frame, and their similarity was evaluated using the DTW-based synchronization score.

As an example, Figure 7 to Figure 11 illustrate the time-series data of joint angles for dance-scene1 to dance-scene5.

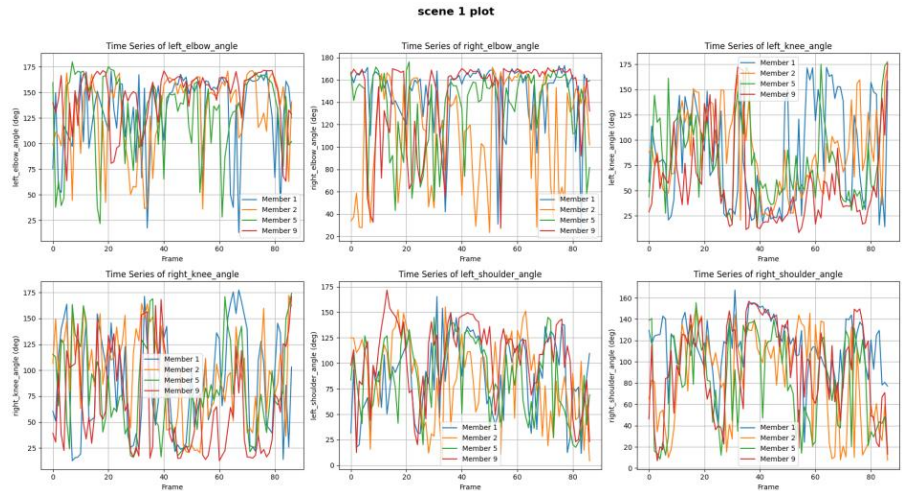

Figure. 7 Time series data for joint angles in dance-scene1

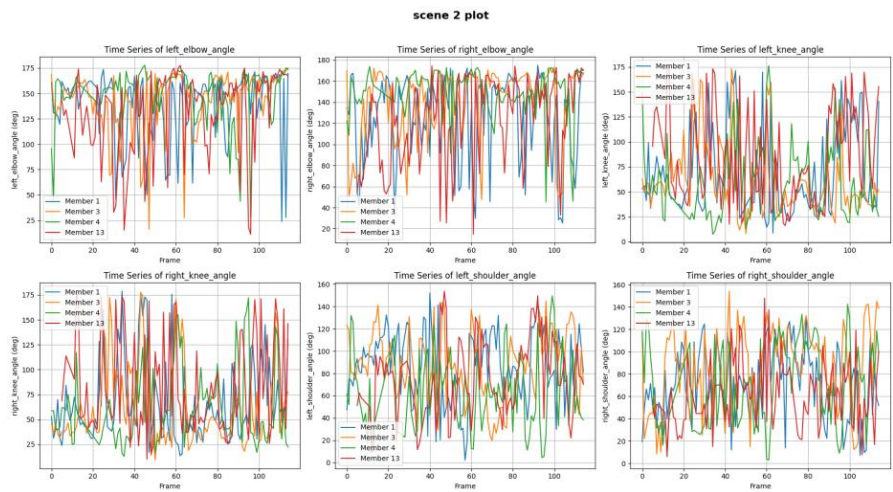

Figure. 8 Time series data for joint angles in dance-scene2

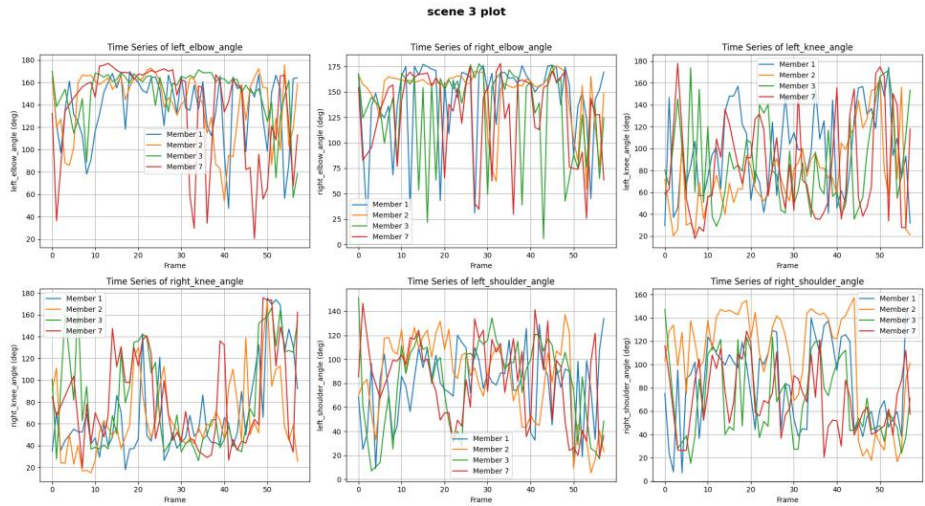
Figure. 9 Time series data for joint angles in dance-scene3

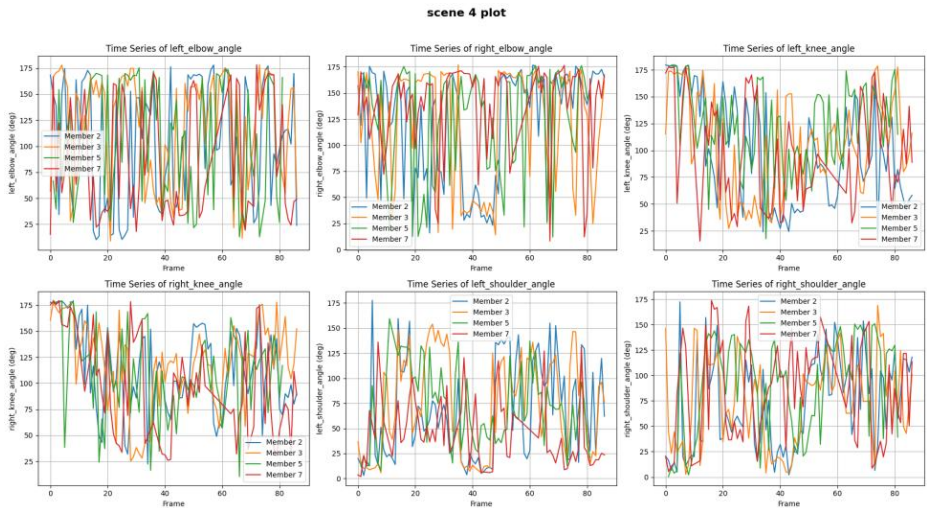
Figure. 10 Time series data for joint angles in dance-scene4

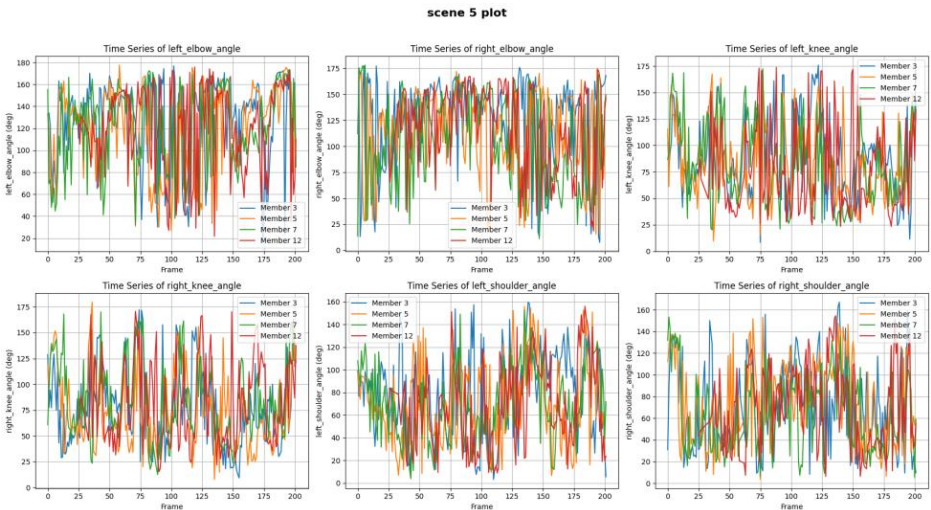
Figure. 11 Time series data for joint angles in dance-scene5

The top row of each figure shows the three time series data for the left elbow, right elbow and left knee, and the bottom row shows the three time series data for the right knee, left shoulder and right shoulder. The horizontal axis shows the number of frames, and the vertical axis shows the joint angle.

These time series data were evaluated using the synchronization rate score calculated by DTW. The synchronization rate scores for each joint in dance-scene1 to dance-scene5 are shown in Tables 5.1 to 5.5.

From Table 5.1 to Table 5.5, the degree of agreement of joint angles was low for all joints.

The highest synchronization score was about 26.66% for the left shoulder in dance-scene1, and the lowest synchronization score was about 3.62% for the left knee in dance-scene5. Similarly, low synchronization scores were calculated for all joints in dance-scenes 6 to 20.

Table 1 DTW Synchronization Score by Distance in dance-scene1

| Feature | Avg_DTW_Distance | Max_DTW_Distance | Synchrony_Rate (%) |
| --- | --- | --- | --- |
| left_elbow_angle | 105.4450115 | 114.9712961 | 8.2857939 |
| right_elbow_angle | 150.1954151 | 190.4899192 | 21.1530900 |
| left_knee_angle | 146.0166300 | 188.2958179 | 22.4535990 |
| right_knee_angle | 156.7016208 | 206.8942211 | 24.2600301 |
| left_shoulder_angle | 123.6863497 | 168.6360410 | 26.6548545 |
| right_shoulder_angle | 109.1126614 | 147.4027513 | 25.9765097 |

Table 2 DTW Synchronization Score by Distance in dance-scene2

| Feature | Avg_DTW_Distance | Max_DTW_Distance | Synchrony_Rate (%) |
| --- | --- | --- | --- |
| left_elbow_angle | 128.5778282 | 153.2900159 | 16.1211985 |
| right_elbow_angle | 165.3208139 | 202.7488778 | 18.4603063 |
| left_knee_angle | 138.1169804 | 147.0899081 | 6.1003014 |
| right_knee_angle | 141.2823264 | 148.3484273 | 4.7631788 |
| left_shoulder_angle | 123.4952205 | 134.7914982 | 8.3805565 |
| right_shoulder_angle | 139.6504933 | 164.5958487 | 15.1555192 |

Table 3 DTW Synchronization Score by Distance in dance-scene3

| Feature | Avg_DTW_Distance | Max_DTW_Distance | Synchrony_Rate (%) |
| --- | --- | --- | --- |
| left_elbow_angle | 99.0104453 | 126.3283114 | 21.6245003 |
| right_elbow_angle | 137.2950897 | 153.5477113 | 10.5847371 |
| left_knee_angle | 100.8461732 | 122.7376138 | 17.8359673 |
| right_knee_angle | 90.4792790 | 103.7377060 | 12.7807212 |
| left_shoulder_angle | 78.6892590 | 90.1215273 | 12.6853912 |
| right_shoulder_angle | 113.1941403 | 139.5545461 | 18.8889625 |

Table 4 DTW Synchronization Score by Distance in dance-scene4

| Feature | Avg_DTW_Distance | Max_DTW_Distance | Synchrony_Rate (%) |
| --- | --- | --- | --- |
| left_elbow_angle | 209.4187469 | 276.5752253 | 24.2814512 |
| right_elbow_angle | 150.6513728 | 162.9006101 | 7.5194545 |
| left_knee_angle | 137.1406883 | 148.1310350 | 7.4193411 |
| right_knee_angle | 119.3873014 | 136.5188158 | 12.5488302 |
| left_shoulder_angle | 150.3665836 | 185.6687145 | 19.0135053 |
| right_shoulder_angle | 137.4373798 | 158.6948234 | 13.3951714 |

Table 5 DTW Synchronization Score by Distance in dance-scene5

| Feature | Avg_DTW_Distance | Max_DTW_Distance | Synchrony_Rate (%) |
|---|---|---|---|
| left_elbow_angle | 191.0266289 | 216.8302451 | 11.9003768 |
| right_elbow_angle | 213.7355668 | 244.6651345 | 12.6415919 |
| left_knee_angle | 126.8864707 | 131.6477893 | 3.6167099 |
| right_knee_angle | 140.7206582 | 149.8874857 | 6.1158058 |
| left_shoulder_angle | 135.3045008 | 154.9657036 | 12.6874543 |
| right_shoulder_angle | 155.3835285 | 173.6835876 | 10.5364354 |

### 4.2.2 Evaluation based on the direction of movement of body parts

Next, for each of the four joints of the four people obtained in 5.2.1, eight directional vectors were created for the following pairs of joints: shoulder to left elbow, left elbow to left wrist, right shoulder to right elbow, right elbow to right wrist, left hip to left knee, left knee to left ankle, right hip to right knee, and right knee → right ankle. The eight directional vectors were evaluated using a cosine similarity score to see how well the directions of the movements matched.

The cosine similarity scores for each joint in dance-scene1 to 5 are shown in Tables 6 to 10. It was confirmed that certain body parts exhibited high synchronicity scores regarding movement direction consistency. In particular, the left knee -> left ankle and right knee -> right ankle consistently showed high values across all scenes of dance-scene1 to dance-scene5, with both parts reaching approximately 90% in dance-scene3. Although some parts, such as the right shoulder -> right elbow (approximately 26%) and the right elbow -> right wrist (approximately 23%) in dance-scene3, showed relatively low values, the overall synchronicity scores were higher than joint angles.

Table 6 Synchronization rate score based on cosine similarity of dance-scene1

| Joint_Pair | Avg_Cosine_similarity(%) |
|---|---|
| Left Shoulder -> Left Elbow | 74.207650 |
| Left Elbow -> Left Wrist | 73.609179 |
| Right Shoulder -> Right Elbow | 60.631295 |
| Right Elbow -> Right Wrist | 76.741989 |
| Left Hip -> Left Knee | 64.306200 |
| Left Knee -> Left Ankle | 86.125718 |
| Right Hip -> Right Knee | 59.663221 |
| Right Knee -> Right Ankle | 86.642127 |

Table 7 Synchronization rate score based on cosine similarity of dance-scene2

| Joint_Pair | Avg_Cosine_similarity |
|---|---|
| Left Shoulder -> Left Elbow | 73.492897 |
| Left Elbow -> Left Wrist | 80.921413 |
| Right Shoulder -> Right Elbow | 53.519351 |
| Right Elbow -> Right Wrist | 79.401441 |
| Left Hip -> Left Knee | 71.428161 |
| Left Knee -> Left Ankle | 76.407420 |
| Right Hip -> Right Knee | 73.726561 |
| Right Knee -> Right Ankle | 81.355261 |

Table 8 Synchronization rate score based on cosine similarity of dance-scene3

| Joint_Pair | Avg_Cosine_similarity |
|---|---|
| Left Shoulder -> Left Elbow | 73.520568 |
| Left Elbow -> Left Wrist | 74.599029 |
| Right Shoulder -> Right Elbow | 25.988357 |
| Right Elbow -> Right Wrist | 22.739364 |
| Left Hip -> Left Knee | 72.448344 |
| Left Knee -> Left Ankle | 90.872652 |
| Right Hip -> Right Knee | 74.606524 |
| Right Knee -> Right Ankle | 89.724648 |

Table 9 Synchronization rate score based on cosine similarity of dance-scene4

| Joint_Pair | Avg_Cosine_similarity |
|---|---|
| Left Shoulder -> Left Elbow | 52.068284 |
| Left Elbow -> Left Wrist | 46.788036 |
| Right Shoulder -> Right Elbow | 47.606353 |
| Right Elbow -> Right Wrist | 58.601642 |
| Left Hip -> Left Knee | 68.836454 |
| Left Knee -> Left Ankle | 70.146110 |
| Right Hip -> Right Knee | 64.665624 |
| Right Knee -> Right Ankle | 75.049364 |

Table 10 Synchronization rate score based on cosine similarity of dance-scene5

| Joint_Pair | Avg_Cosine_similarity |
|---|---|
| Left Shoulder -> Left Elbow | 54.839805 |
| Left Elbow -> Left Wrist | 53.423290 |
| Right Shoulder -> Right Elbow | 56.672473 |
| Right Elbow -> Right Wrist | 56.359274 |
| Left Hip -> Left Knee | 78.939313 |
| Left Knee -> Left Ankle | 83.433204 |
| Right Hip -> Right Knee | 77.986581 |
| Right Knee -> Right Ankle | 82.715908 |

### 4.2.3 Evaluation based on jumping movements

From Tables 11 to 15, it was confirmed that certain body parts exhibited high synchronicity scores regarding movement direction consistency. In particular, the left knee -> left ankle and right knee -> right ankle consistently showed high values across all scenes of dance-scene1 to dance-scene5, with both parts reaching approximately 90% in dance-scene3. Although some parts, such as the right shoulder -> right elbow (approximately 26%) and the right elbow -> right wrist (approximately 23%) in dance-scene3, showed relatively low values, the overall synchronicity scores were higher than joint angles.

Table 11 Synchronization score for jump-scene1

| Feature | Head_Position_Synchrony | Foot_Position_Synchrony |
|---|---|---|
| jump_motion | 96.417604 | 91.452321 |

Table 12 Synchronization score for jump-scene2

| Feature | Head_Position_Synchrony | Foot_Position_Synchrony |
|---|---|---|
| jump_motion | 92.316904 | 95.980394 |

Table 13 Synchronization score for jump-scene3

| Feature | Head_Position_Synchrony | Foot_Position_Synchrony |
|---|---|---|
| jump_motion | 91.794637 | 92.545314 |

Table 14 Synchronization score for jump-scene4

| Feature | Head_Position_Synchrony | Foot_Position_Synchrony |
|---|---|---|
| jump_motion | 98.456101 | 98.237707 |

Table 15 Synchronization score for jump-scene5

| Feature | Head_Position_Synchrony | Foot_Position_Synchrony |
|---|---|---|
| jump_motion | 96.166843 | 98.447036 |

### 4.2.4 Evaluation based on crouching movements

The consistency of crouching motions in down-scene1 to down-scene5 was evaluated based on head height during the crouching motion. Using the same method as in the jump motion analysis, synchronicity scores were calculated for the head heights of the four participants in the three-dimensional space.

Tables 16 to 20 present the synchronicity scores for head height in down-scene1 to down-scene5. From these tables, it was confirmed that high synchronicity scores were obtained for head height during the crouching motion. In all scenes except for down-scene2, values exceeding 90% were observed. Even in down-scene2, the synchronicity score was approximately 89.7%, which is close to 90%.

Table 16 Synchronization score for down-scene1

| Feature | Head_Position_Synchrony |
|---|---|
| down_motion | 93.321806 |

Table 17 Synchronization score for down-scene2

| Feature | Head_Position_Synchrony |
|---|---|
| down_motion | 89.660313 |

Table 18 Synchronization score for down-scene3

| Feature | Head_Position_Synchrony |
|---|---|
| down_motion | 93.925663 |

Table 19 Synchronization score for down-scene4

| Feature | Head_Position_Synchrony |
|---|---|
| down_motion | 94.053916 |

Table 20 Synchronization score for down-scene5

| Feature | Head_Position_Synchrony |
|---|---|
| down_motion | 96.183151 |

# 5  Discussion and Conclusion

In this study, we investigated the factors contributing to SEVENTEEN's reputation for having a "high synchronization rate." We conducted member-specific tracking and 2D skeleton detection for dance scenes categorized as "Kal-gunmu" (perfectly synchronized group choreography) using YOLOv8. Additionally, we combined this with 3D pose estimation using MediaPipe to evaluate the consistency of dance movements in 3D space.

To analyze joint angles, we calculated synchronization scores using Dynamic Time Warping (DTW) and Dynamic Barycenter Averaging (DBA). For body part movement directions, we calculated synchronization scores using cosine similarity. Furthermore, synchronization scores were derived from the absolute difference between the average positions of the ankles and head for jumping movements. Similarly, synchronization scores were calculated for crouching movements based on the absolute difference in head position.

As a result, we obtained the following two key findings:
1. Joint Angles and Body Part Movement Directions
    - The synchronization scores for joint angles calculated using DTW were low for all joints across all scenes, indicating variability in the magnitude of joint angles during dance movements.
    - In contrast, the synchronization scores for body part movement directions, calculated using cosine similarity, showed high values overall, suggesting that the movement directions of body parts were well-aligned.
    - These findings indicate that SEVENTEEN's high synchronization rate is more attributable to the consistency of body part movement directions rather than the uniformity of joint angles.
2. Jumping and Crouching Movements
    - The synchronization scores for ankle and head height during jumping movements were high.
    - Similarly, the synchronization scores for head height during crouching movements were also high.
    - These results suggest that SEVENTEEN maintains synchronization by aligning the peak heights of the ankles and head during jumps and the lowest head positions during crouching, regardless of differences in individual body proportions.

Our future research should explore alternative evaluation methods. For example, machine learning techniques could extract features and classify choreography, allowing for a more detailed analysis of which types of movements increase or decrease synchronization.

Additionally, this study focused on Kal-gunmu, where all members perform identical choreography with high synchronization. However, further analysis is needed to compare Kal-gunmu with other dance styles using the same evaluation methods. This comparison would provide a deeper understanding of synchronization levels in different types of dance movements.

## Acknowledgements

I am grateful to SEVENTEEN, BUMZU and all the wonderful songs they produced.

## Author contributions statement

A.S proposed a method and conducted experiments under supervise of H.H. A.S and H.H analysed the results. After H.H verified the results, all the authors reviewed the manuscript.

## Additional information

This manuscript is an edited and English-translated part of the Atsushi Shimojo's graduation thesis from the Department of Information Engineering, Faculty of Engineering, Ibaraki University, completed in the academic year 2024. The author enjoys playing the guitar as a hobby and has formed a cover band of Thee Michelle Gun Elephant with university friends. For the author, SEVENTEEN's music was something to be enjoyed purely through sound, with little interest in their visuals. While the author sometimes plays SEVENTEEN's songs by ear on the guitar, the most important aspect is the way their compositions perfectly align with the moments when one thinks, "It would feel great if the music progressed this way" while playing. In that sense, the author might not be a CARAT but simply a Woozi Pen. However, for many CARATs, it is impossible to talk about the appeal of SEVENTEEN without mentioning their dance performances. Through this study, the author feels they have gained some understanding of that perspective. Perhaps, just as the author finds pleasure in satisfying chord progressions, CARATs experience a similar feeling when SEVENTEEN's performances synchronize at just the right moments.

# Appendix
# List of analysis videos

- dance-scene1：[Dance Practice] SEVENTEEN - Adore U - Fixed Cam Ver.
  https://www.youtube.com/watch?v=wPYLRdLZ5ts

- dance-scene2：[Dance Practice] SEVENTEEN - Adore U - Fixed Cam Ver.
  https://www.youtube.com/watch?v=wPYLRdLZ5ts

- dance-scene3：[Dance Practice] SEVENTEEN - Adore U - Fixed Cam Ver.
  https://www.youtube.com/watch?v=wPYLRdLZ5ts

- dance-scene4：[Dance Practice] SEVENTEEN - Adore U - Fixed Cam Ver.
  https://www.youtube.com/watch?v=wPYLRdLZ5ts

- dance-scene5：[SPECIAL VIDEO] SEVENTEEN - VERY NICE DANCE PRACTICE ver.
  https://www.youtube.com/watch?v=A1gJQpMSkEU

- dance-scene6：[SPECIAL VIDEO] SEVENTEEN - VERY NICE DANCE PRACTICE ver.
  https://www.youtube.com/watch?v=A1gJQpMSkEU

- dance-scene7：[SPECIAL VIDEO] SEVENTEEN - VERY NICE DANCE PRACTICE ver.
  https://www.youtube.com/watch?v=A1gJQpMSkEU

- dance-scene8：[SPECIAL VIDEO] SEVENTEEN - VERY NICE DANCE PRACTICE ver.
  https://www.youtube.com/watch?v=A1gJQpMSkEU

- dance-scene9：[Choreography Video] SEVENTEEN- Don't Wanna Cry Front Ver.
  https://www.youtube.com/watch?v=aJ55DJE4Os4

- dance-scene10：[Choreography Video] SEVENTEEN- Don't Wanna Cry Front Ver.
  https://www.youtube.com/watch?v=aJ55DJE4Os4

- dance-scene11：[Choreography Video] SEVENTEEN - CLAP
  https://www.youtube.com/watch?v=K qlkKYL-54

- dance-scene12：[Choreography Video] SEVENTEEN - CLAP
  https://www.youtube.com/watch?v=K qlkKYL-54

- dance-scene13：[Choreography Video] SEVENTEEN - Getting Closer
  https://www.youtube.com/watch?v=oJWYUOTGrMA

- dance-scene14：[Choreography Video] SEVENTEEN – Home
  https://www.youtube.com/watch?v=8EfdpvSLE-U

- dance-scene15：[Choreography Video]SEVENTEEN - Happy Ending
  https://www.youtube.com/watch?v=CJZ8xUA2aYk

- dance-scene16：[Choreography Video] SEVENTEEN - HOME;RUN
  https://www.youtube.com/watch?v=IAuTjzenUMg

- dance-scene17：[Choreography Video] SEVENTEEN - Ready to love
  https://www.youtube.com/watch?v=YxWowt5Oc9Y

- dance-scene18：[Choreography Video] SEVENTEEN – Anyone
  https://www.youtube.com/watch?v=YOv1EbQyh1U

- dance-scene19：[Choreography Video] SEVENTEEN – Anyone
  https://www.youtube.com/watch?v=YOv1EbQyh1U

- dance-scene20：[Choreography Video] SEVENTEEN - My My
  https://www.youtube.com/watch?v=w4RdpL2Fp8o

- jump-scene1：[Choreography Video] SEVENTEEN- Don't Wanna Cry Front Ver.
  https://www.youtube.com/watch?v=aJ55DJE4Os4

- jump-scene2：[SPECIAL VIDEO] SEVENTEEN - VERY NICE DANCE PRACTICE ver.
  https://www.youtube.com/watch?v=A1gJQpMSkEU

- jump-scene3：[Choreography Video] SEVENTEEN - Ready to love
  https://www.youtube.com/watch?v=YxWowt5Oc9Y

- jump-scene4：[Choreography Video] SEVENTEEN - Ready to love
  https://www.youtube.com/watch?v=YxWowt5Oc9Y

- jump-scene5：[Choreography Video] SEVENTEEN – HIT
  https://www.youtube.com/watch?v=loPVNjGkrz4

- down-scene1：[Choreography Video] SEVENTEEN - CLAP
  https://www.youtube.com/watch?v=K qlkKYL-54

- down-scene2：[Choreography Video] SEVENTEEN - Getting Closer
  https://www.youtube.com/watch?v=oJWYUOTGrMA

- down-scene3：[Choreography Video]SEVENTEEN - Happy Ending
  https://www.youtube.com/watch?v=CJZ8xUA2aYk

- down-scene4：[Choreography Video] SEVENTEEN - My My
  https://www.youtube.com/watch?v=w4RdpL2Fp8o

- down-scene5：[Choreography Video] SEVENTEEN- Don't Wanna Cry Front Ver.
  https://www.youtube.com/watch?v=aJ55DJE4Os4